\documentclass[lettersize,journal]{IEEEtran}
\usepackage{amsmath,amsfonts}
\usepackage{algorithmic}
\usepackage{algorithm}
\usepackage{array}
\usepackage[caption=false,font=normalsize,labelfont=sf,textfont=sf]{subfig}
\usepackage{textcomp}
\usepackage{stfloats}
\usepackage{url}
\usepackage{verbatim}
\usepackage{graphicx}
\usepackage{cite}
\usepackage{color}
\usepackage{tikz}
\usepackage{comment}

\usepackage{times}

\usepackage{graphicx}
\usepackage{booktabs}
\usepackage{multicol}

\usepackage{graphicx}
\usepackage{array}

\usepackage{multirow}
\usepackage{algorithm}
\usepackage{algorithmic}
\usepackage{
caption}
\usepackage{epsfig}
\usepackage{amsthm, amsfonts, amssymb, gensymb}
\usepackage{booktabs}
\usepackage{makecell}
\usepackage{subcaption}
\usepackage{setspace}
\usepackage{textcomp}
\usepackage{wrapfig}
\usepackage{marvosym}

\usepackage{verbatim}

\usepackage[nolist,nohyperlinks]{acronym}
\acrodef{SG-ZSL}{Sentinel-Guided Zero-Shot Learning}

\newcommand{\ie}{\textit{i}.\textit{e}.}

\begin{document}

\title{Sentinel-Guided Zero-Shot Learning: A Collaborative Paradigm without Real Data Exposure}

\author{Fan Wan, Xingyu Miao, Haoran Duan, Jingjing Deng, Rui Gao, Yang Long \Letter,~\IEEEmembership{Senior Member~IEEE}
\thanks{Fan Wan, Xingyu Miao, Haoran Duan, Jingjing Deng, Yang Long are with the Department of Computer Science, Durham University (E-mail:\{fan.wan; xingyu.miao; haoran.duan; jingjing.deng; yang.long\}@durham.ac.uk)}
\thanks{Rui Gao is with School of Electronic and Information Engineering, Xi'an Jiaotong University, Xi'an, China (E-mail:\{gaorui1013@stu.xjtu.edu.cn\})}
\thanks{Fan Wan, Rui Gao and Haoran Duan are equal contributions.}
\thanks{Yang Long is the corresponding author \Letter.}
}

\markboth{IEEE Transactions on Circuits and Systems for Video Technology}
{Shell \MakeLowercase{\textit{et al.}}: A Sample Article Using IEEEtran.cls for IEEE Journals}
\maketitle

\begin{abstract}
With increasing concerns over data privacy and model copyrights, especially in the context of collaborations between AI service providers and data owners, an innovative \ac{SG-ZSL} paradigm is proposed in this work. 
\ac{SG-ZSL} is designed to foster efficient collaboration without the need to exchange models or sensitive data. It consists of a teacher model, a student model and a generator that links both model entities.  The teacher model serves as a sentinel on behalf of the data owner, replacing real data, to guide the student model at the AI service provider's end during training.  Considering the disparity of knowledge space between the teacher and student, we introduce two variants of the teacher model: the omniscient and the quasi-omniscient teachers. Under these teachers' guidance, the student model seeks to match the teacher model's performance and explores domains that the teacher has not covered. To trade-off between privacy and performance, we further introduce two distinct security-level training protocols: white-box and black-box, enhancing the paradigm's adaptability. Despite the inherent challenges of real data absence in the \ac{SG-ZSL} paradigm, it consistently outperforms in ZSL and GZSL tasks, notably in the white-box protocol. Our comprehensive evaluation further attests to its robustness and efficiency across various setups, including stringent black-box training protocol.

\end{abstract}

\begin{IEEEkeywords}
Data-Free Knowledge Transfer, Privacy Protection, Zero-Shot Learning.

\end{IEEEkeywords}

\acresetall


\section{Introduction}
The profound advancements in deep learning can be largely attributed to the evolution of high-performance computing and the proliferation of extensive multimodal datasets. At the heart of deep learning lies the ability of pre-trained models, such as ResNet \cite{he2016deep} for visual features and BERT \cite{devlin2018bert} for semantic nuances, to distill empirical knowledge from vast datasets. Through these models, intricate patterns and relationships within expansive data terrains are effectively discerned, proving pivotal for addressing intricate real-world challenges.
Nonetheless, the sharing of data among institutions has increasingly been met with complexities and apprehensions. The broader public's concerns regarding data ownership, copyright implications, the financial burden of large-scale annotations, and restricted access to domain-specific data have hindered the progression of interdisciplinary and intercultural deep learning models. 
Furthermore, the hesitation to share models publicly often comes from concerns about intellectual property, misuse, and protecting private and sensitive information.
Balancing the desire for openness with these issues is an ongoing struggle for researchers.

\begin{figure}[t]
    \centering
    \includegraphics[width=\linewidth]{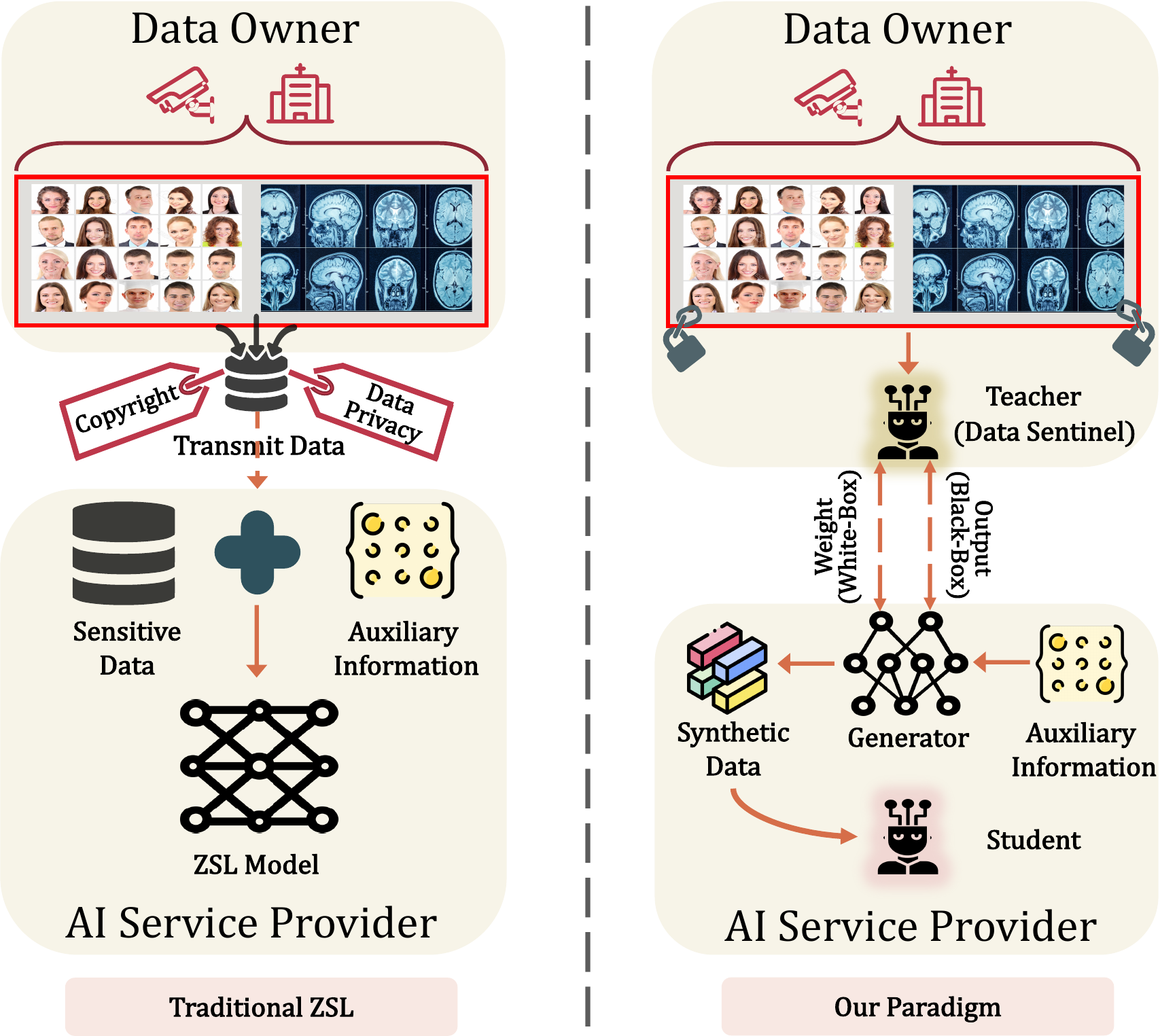}
    \caption{In traditional ZSL approaches, real data is necessitated to establish the visual-semantic association. Conversely, SG-ZSL introduces a teacher model, which acts as a data sentinel, enabling the execution of ZSL tasks without the need for direct access to real data.
    }
    \label{zsl-AZSL}
\end{figure}

As depicted in Fig.\ref{zsl-AZSL}, datasets, which may encompass sensitive entities such as personal healthcare records and facial images, have necessitated substantial financial and temporal investments from data proprietors. Stringent regulations, epitomized by Europe's GDPR and the California Consumer Privacy Act (CCPA), have been instituted to safeguard personal data and uphold user privacy. As a result, the acquisition, transmission, and dissemination of such data have become increasingly intricate and laden with challenges.
Federated Learning (FL) \cite{konevcny2016Federated}, as a pioneering privacy-preserving paradigm, offers a decentralized approach to model training. This method facilitates the training of a global shared model across multiple parties through the mere exchange of model updates without centralizing data, which largely preserves data privacy. Nonetheless, issues coexist, including ambiguities in model ownership, potential disclosure of proprietary information, uncertainties in intellectual property rights, and threats stemming from unauthorized model deployment.
The fact that many AI service providers are unwilling to disclose the training datasets and internal parameters of Large Language Models is an illustration of these dilemmas.
Given the mutual dependencies between AI service providers (i.e., AI companies) and data owners (i.e., research institutions and hospitals), direct data or model sharing poses potential infringements on copyright and privacy mandates. 
This situation underscores the critical demand for inventive approaches that enable collaboration while safeguarding sensitive information and adhering to copyright laws.
Against this backdrop of challenges, our study embarks on an exploration within a strict framework where neither real data nor models are disclosed to the public.
We endeavor to elucidate a collaborative learning paradigm necessitating no real data exchange, wherein knowledge transfer transpires through a teacher-student distillation facilitated by a data-generative mechanism.

Zero-shot learning (ZSL), an emerging machine learning paradigm, demonstrates considerable potential in addressing data-free challenges, particularly in scenarios where deep transfer extends beyond the seen classes present in the training dataset. By establishing a robust visual-semantic linkage through auxiliary information, such as attributes or word embeddings, ZSL empowers models to recognize unseen classes without prior exposure to relevant data. Nevertheless, conventional ZSL models predominantly depend on actual data from either seen or unseen categories. In adapting pre-trained models to novel task domains, an implicit assumption is made regarding the presence of a significant volume of labeled seen class or unlabeled unseen class data to build the visual-semantic connection. This assumption, however, is often contradicted to the restrictive nature of data sharing across varied institutions and countries.

In this paper, we introduce \ac{SG-ZSL}, as shown in Fig. \ref{zsl-AZSL}. The proposed approach markedly diverges from traditional ZSL, aiming to prevent the leakage of sensitive data while still enabling effective training of AI models. Within this framework, the teacher model, pre-trained on real data, assumes the dual responsibility of data protection and guidance for the ZSL model located at the AI Service Provider's end, thereby ensuring model training without direct exposure to the original data from the Data Owner. 
To reconcile privacy concerns with performance objectives, two distinct SG-ZSL protocols are presented: a `black-box' version, which solely conveys the output classification scores from the teacher model and a `white-box' variant that shares both model weights and classification scores. These protocols are tailored to meet the diverse needs of data owners and AI service providers concerning model efficacy and data security. To bolster defenses against potential threats, Differential Privacy (DP) \cite{dwork2008differential} is integrated into our teacher model's training process. The direct transfer of the teacher model to the AI Service Provider is prohibited; instead, the latter must submit specific requests to receive training guidance from the teacher model. This mechanism allows the Data Owner to exert control over data security by regulating request frequency. Recognizing potential disparities in the knowledge domains of the teacher and student models, our teacher models are further categorized into omniscient and quasi-omniscient types, distinguished primarily based on their exposure to unseen categories during the pre-training phase.

In summary, our contributions encompass:
\begin{itemize}
\item We introduce Sentinel-Guided Zero-Shot Learning, a novel paradigm for Zero-Shot classification without real data access, addressing crucial data privacy and model copyright issues.
\item To meet the multifaceted needs in terms of privacy preservation and performance optimization, we formulate two distinct training protocols: white-box and black-box. Additionally, we analyze teacher models under omniscient and quasi-omniscient scenarios within the knowledge space, enhancing our paradigm's adaptability.
\item We showcase experimental results for both conventional and generalized ZSL tasks in two scenarios. Despite the lack of data sharing during training, the SG-ZSL model yields promising performance, highlighting our approach's viability.
\end{itemize}

\renewcommand\arraystretch{5}

\begin{table*}[!t]

\caption{The distinctions between SG-ZSL and traditional ZSL settings are delineated in the table. Herein, `S' and `U' denote the seen and unseen classes, respectively. `$\mathcal X$' signifies visual features, while `$\tilde{\mathcal{X}}$' pertains to generated features. The semantics of the seen and unseen classes are represented by `$A_s$' and `$A_u$', respectively. The red `X' symbolizes sensitive real data. The ZSL model is denoted by `$\theta$', whereas `$\theta_T$' corresponds to the pre-trained teacher model specific to the SG-ZSL task. `$\theta_U$' can be associated with either the conventional ZSL model or the SG-ZSL model. It should be noted that the SG-ZSL model is constructed under the guidance of the teacher model, effectively eliminating the need for sharing actual data.}

\resizebox{\textwidth}{!}
{
\begin{tabular}{cccc}
\hline\hline\\[-88pt]
                    & \textbf{IZSL}    & \textbf{TZSL}& \textbf{SG-ZSL} \\[-12pt]\hline

\specialrule{0em}{-10pt}{-18pt}

\makecell[c]{\textbf{Accessible Data}}              &  $\mathcal X_s$       & $\mathcal X_s \cup \mathcal X_u$   &  \textbf{0}   \\

\specialrule{0em}{-10pt}{-30pt}

\makecell[c]{\textbf{Accessible Weights}}   &  $\theta_S$    &    $\theta_{S+U}$      & \textbf{$\theta_{T}$} $/$ \textbf{0}   \\

\specialrule{0em}{-3pt}{-20pt}

\makecell[c]{\textbf{Paradigm}}                &    \begin{minipage}[b]{0.5\columnwidth}
		\centering
		\raisebox{-.6\height}{\includegraphics[width=0.98\linewidth]{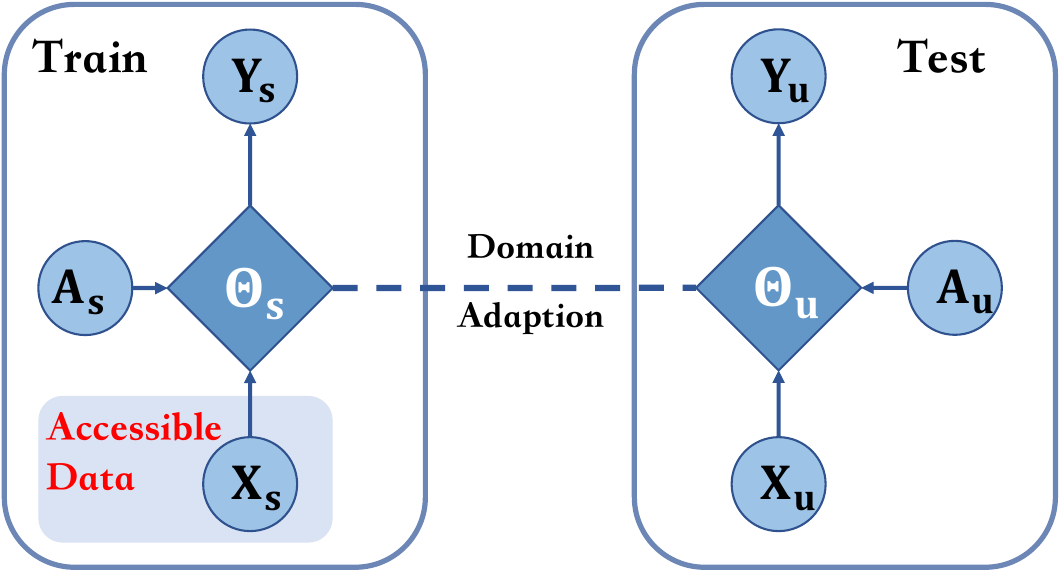}}
	\end{minipage}     &  \begin{minipage}[b]{0.5\columnwidth}
		\centering
		\raisebox{-.6\height}{\includegraphics[width=0.98\linewidth]{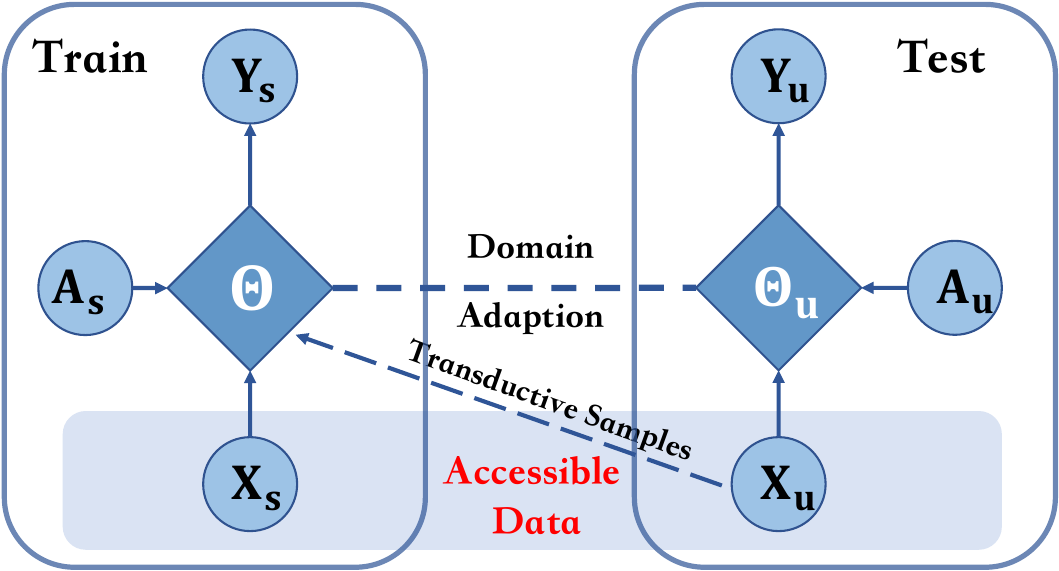}}
	\end{minipage} &   \begin{minipage}[b]{0.5\columnwidth}
		\centering
		\raisebox{-.6\height}{\includegraphics[width=0.98\linewidth]{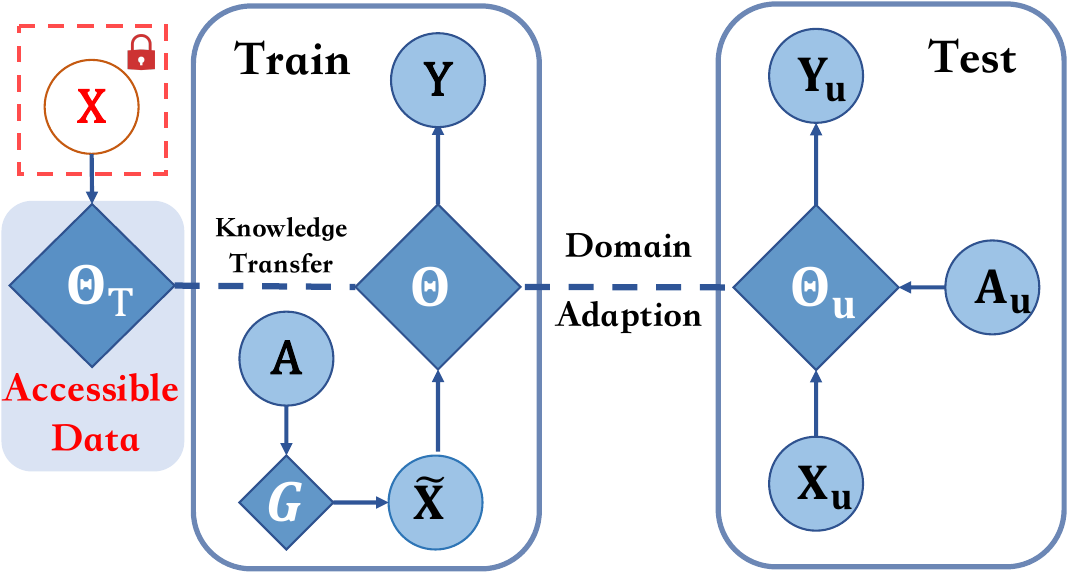}}
	\end{minipage}     \\

\specialrule{0em}{5pt}{0pt}
\hline\hline                      
\end{tabular}
\label{formulation} 
}
\end{table*}

\section{Related Work}
The realm of machine learning has recently experienced a significant shift towards prioritizing data privacy, particularly when handling sensitive information across diverse domains. Federated Learning \cite{konevcny2016} has been recognized as a formidable framework, designed to mitigate potential data leakage by decentralizing the training process. Recent advancements in this domain have been characterized by the exploration of various architectures and optimization strategies, all aimed at enhancing model performance without sacrificing data privacy. For example, studies \cite{zhang2023spatial,reisizadeh2020fedpaq, wan2023community,ren2024fedboosting} have been dedicated to optimizing communication efficiency in federated learning setups, while research such as \cite{wang2019adaptive, yu2021toward} has delved into the application of federated learning in edge computing, ensuring data privacy at its source.

Differential Privacy \cite{dwork2008differential} has been seamlessly integrated into numerous machine learning paradigms to bolster data privacy. Recent contributions, including \cite{guo2021topology, fu2023gc, 10006830}, have investigated the fusion of DP with deep learning, ensuring that while models remain proficient, the privacy of their training data remains uncompromised. For example, Guo \textit{et al.}\cite{guo2021topology} developed `TOP-DP', a topology-aware differential privacy approach for decentralized image classification systems, which innovatively utilizes decentralized communication topologies to enhance privacy protection while achieving an improved balance between model usability and data privacy. 

Knowledge Distillation \cite{xu2020computation}, on the other hand, has emerged as a pivotal strategy for protecting intricate teacher models by training a streamlined student model, thereby thwarting potential adversarial attacks. Recent endeavors, such as \cite{9248028, 10148979, beyer2022knowledge, 9461003}, have showcased the versatility of knowledge distillation across domains of computer vision. For example, Zhang \textit{et al.}\cite{9461003} introduced an evolutionary knowledge distillation approach, where an adaptive, online-evolving teacher model continuously transfers intermediate knowledge to a student network, significantly enhancing learning effectiveness, especially in low-resolution and few-sample scenarios.

It is imperative to note, however, that both Federated Learning and Knowledge Distillation are predominantly confined to supervised learning. This confines their utility in scenarios necessitating the recognition and categorization of previously unseen data categories, a domain where Zero-Shot Learning protocols excel. ZSL, with its prowess in recognizing unseen classes by establishing semantic relationships, transcends the limitations inherent to the supervised nature of both Federated Learning and Knowledge Distillation.

In this work, an innovative SG-ZSL paradigm is introduced. This paradigm, distinct in its data-free knowledge transfer, is adept at addressing unseen data categories, especially in contexts where data sensitivity and privacy are paramount. The incorporation of DP within the teacher model further enhances data privacy, ensuring that the traditional ZSL generalization properties to unseen classes are preserved without additional training, all while safeguarding data and model privacy.

Zero-Shot Learning \cite{zhang2018triple, larochelle2008zero, vyas2020leveraging, gao2023privacy, zhang2019probabilistic} is predicated on recognizing unseen classes by establishing connections between seen and unseen classes through semantic information, such as attributes \cite{jayaraman2014zero, li2023hierarchical,long2017towards,guo2018zero}, word embeddings \cite{zhang2016zero} and predefined similes \cite{long2017describing,zhang2019zero}. Numerous studies \cite{qin2016beyond, kodirov2017semantic, felix2018multi} have been dedicated to mapping from visual to semantic space, while others \cite{wang2023deconfounding, gao2020zero, xian2018feature,cheng2023hybrid} focus on generating unseen class data to mitigate data scarcity issues. Effective spaces for visual and semantic embedding have been investigated in \cite{akata2013label, akata2015evaluation, tian2019aligned,liu2018zero,liu2022zero}. Depending on the utilization of unseen data during training, ZSL methods can be categorized into inductive \cite{long2017zero, romera2015embarrassingly} and transductive settings \cite{song2018transductive, fu2015transductive}. As for the test phase, conventional ZSL methods \cite{akata2013label, norouzi2013zero} operate under the assumption that test data originates exclusively from unseen classes, while Generalized ZSL (GZSL) \cite{chao2016empirical, min2020domain, hu2023domain} aims to classify both seen and unseen data into their respective classes.

The distinctions between SG-ZSL and traditional ZSL settings are elucidated in Table \ref{formulation}. In terms of data access during training, IZSL and Transductive ZSL (TZSL) access labeled seen data and data from both seen and unseen classes, respectively. In contrast, the SG-ZSL setting operates without direct data access, relying solely on a teacher model, trained on sensitive real data, for guidance (as indicated by the red ‘X' in Table \ref{formulation}). Concerning model security, weight accessibility refers to the accessibility of weights trained on real data. While ZSL models in both inductive and transductive settings possess accessible weights, the SG-ZSL paradigm introduces a teacher model pre-trained on real data. In assessing teacher weight privacy, we introduce the black-box and white-box protocols. In the white-box protocol, teacher weights are accessible for guidance during SG-ZSL model training, whereas the black-box protocol restricts weight sharing, thereby preserving the privacy of both data and model weights.


\section{Methodology}

As depicted in Fig. \ref{zsl-AZSL}, in scenarios where the Data Owner's sensitive data is inaccessible yet a collaboration with the AI Service Provider is sought to leverage the data's value, the proposed SG-ZSL paradigm emerges as a solution. The Data Owner employs a teacher model, serving as a data sentinel, which guides the AI Service Provider's models in training classifiers without real data access. Recognizing the balance between privacy preservation and performance optimization, two distinct training protocols with varying security levels, namely the white-box and black-box protocols, are introduced to enhance the paradigm's adaptability.

\subsection{\textbf{Problem Definition}}
The SG-ZSL paradigm fosters collaboration between the Data Owner, housing a teacher model, and the AI Service Provider, hosting a student model and a generator. The teacher model, represented as $\mathcal{F}_{\theta_{T}}: \mathcal{X}\rightarrow\mathcal{Y}$, serves as a data sentinel. Central to the SG-ZSL paradigm is the utilization of the teacher model at the Data Owner's end to direct the training of the student model at the AI Service Provider's end. This objective is achieved through synthetic data generated by the generator $\mathcal{F}_{\theta_{G}}$, with the aim of enabling the student model to match the teacher's performance or explore domain not covered by the teacher without the transmission of real data. The objective function is given by:

\begin{equation}
\setlength{\abovedisplayskip}{5pt} 
\setlength{\belowdisplayskip}{5pt}
    \ell\left( \mathcal{F}_{\{\theta_{S},\theta_{G}\}}\left(\tilde{x}\right),\mathcal{F}_{\theta_{T}}\left(\tilde{x}\right) \right), 
    \label{E-1}
\end{equation}

where $\ell$ denotes the objective function guided by the teacher, and $\tilde{x}\in\tilde{\mathcal{X}}$ signifies the data generated by the generator, ensuring no real data access.

\begin{figure}[!bt]
    \centering
    \includegraphics[width=\linewidth]{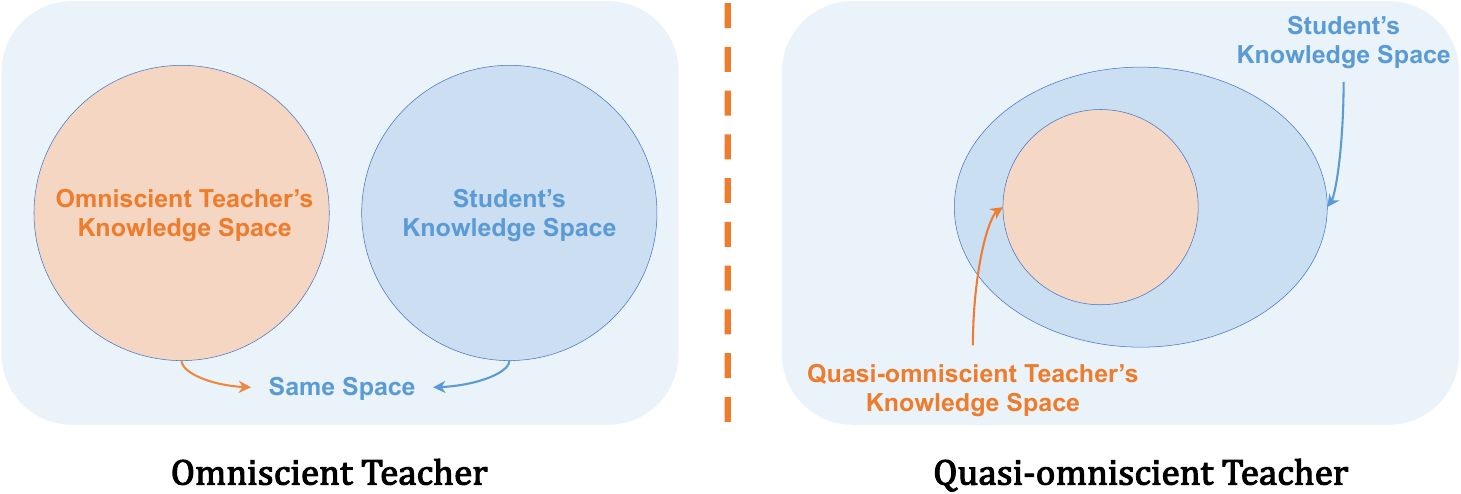}
    \caption{Differences between the Omniscient and the Quasi-omniscient teacher.
    }
    \label{teacher-type}
\end{figure}
\subsection{\textbf{Data Sentinel at the Data Owner's End}}
\subsubsection{\textbf{Omniscient and Quasi-omniscient Teachers}}
Given the potential inconsistency between the teacher's data categories and the student model's objective categories, there may be unseen class data absent in the teacher's domain but essential for the student model. Thus, teacher models are further categorized into omniscient and quasi-omniscient types as shown in Fig.\ref{teacher-type}. The omniscient model encompasses all categories, covering both seen and unseen class data, while the quasi-omniscient model is limited to seen class data.

Here, we define the seen class as $\mathcal S = \{ \left ( x_s, a_s, y_s \right ) \mid$  $ x_s \in \mathcal X_s, a_s \in \mathcal A, y_s \in \mathcal Y_s \}$, where $x_s \in \mathbb{R}^{d_x}$ denotes the  $d_x$-dimensional visual feature in the set of seen class features, $a_s \in \mathbb{R}^{d_a}$ denotes the $d_a$-dimensional auxiliary class-level semantic embedding, and $\mathcal Y_s$ stands for the set of labels for seen classes. Unseen classes are defined as $\mathcal U = \{ \left ( x_u,a_u,y_u \right ) \mid$ $ x_u \in \mathcal X_u, a_u \in \mathcal A, y_u \in \mathcal Y_u\}$, where $x_u$ represents the unseen class features, $a_u$ denotes the semantic embedding of unseen classes and $y_u$ denotes the unseen class labels. The seen and unseen classes are disjoint, \ie, $\mathcal Y_s \cap \mathcal Y_u = \emptyset$.

In the SG-ZSL paradigm, a key constraint is the inaccessibility of both seen and unseen real features at the Data Owner's end during the student model and generator training at the AI Service Provider's end. The available information for the AI service provider is represented as $\mathcal T_r= \{ \left ( a, y \right ) \mid a \in \mathcal A, y \in \mathcal Y \}$, indicating only semantic embeddings $a$ and class labels $y$ are available during training. Additionally, a teacher model, pre-trained on real data, is provided to guide the training of the student model and generator. Depending on the teacher model type, different teacher objectives are considered.

\subsubsection{\textbf{Teacher Objectives}}
The teacher models guide the student model in mastering various ZSL tasks. For the CZSL task, the student model's objective is to classify test images, represented by $f_{ZSL}:\mathcal X_{u} \rightarrow \mathcal Y_u$. For the GZSL task, the student model aims to recognize test images, denoted by $f_{GZSL}:\mathcal X \rightarrow \mathcal Y$.

\subsubsection{\textbf{Incorporating DP in Teacher Model Training}}

To bolster the protection of sensitive data at the Data Owner's end, differential privacy techniques are seamlessly integrated into the teacher model's training process.

Differential privacy stands as a preeminent mechanism for ensuring data and model security. 
Denote an algorithm with the diﬀerential privacy property by $M(.)$.
The algorithm is randomized to make it difficult to have access to the privacy information of the input data.
The formal definition of DP is provided below:

\noindent Definition 1 \cite{dwork2008differential}. Given a pair of neighboring datasets $D$ and $D'$, for every set of outcomes $S$, a mechanism $M$ satisfies DP if the following holds:
\begin{equation}
\mathbb{P}(M(D)\in S)\leq e^{\varepsilon}\cdot \mathbb{P}(M(D')\in S)+\delta     
\end{equation}
Here, M(D) and M(D') represent the algorithm's outputs for input datasets $D$ and $D'$, respectively, and $\mathbb{P}$ captures the algorithm's inherent noise randomness. Both $\varepsilon$ (privacy budget) and $\delta$ (failure probability) influence the privacy strength: smaller values of $\varepsilon$ and $\delta$ ensure enhanced privacy. In the realm of deep learning, DP is typically realized by introducing the subsampled Gaussian mechanism to safeguard the minibatch gradients during the training process \cite{bu2022automatic,dwork2014algorithmic,zhang2021feddpgan}. The distinction between deep learning with DP and conventional deep learning hinges on the private release of the gradient. The Gaussian mechanism is defined as:

\noindent Definition 2 (Gaussian Mechanism) \cite{dwork2014algorithmic}. Let $\Delta f$ be the sensitivity of function $f$, defined as $\Delta f =\mathop{max}\limits_{D,D'} \left \| f(D)-f(D') \right \|_{2}$. The Gaussian Mechanism, $\hat{f}(D)=f(D)+\sigma \Delta f\cdot \mathcal{N}(0,\mathcal{I})$, is deemed ($\varepsilon$, $\delta$)-differentially private for specific values of $\varepsilon$ and $\delta$ contingent on $\sigma$.

During our teacher models' training, random noise is introduced to perturb the original data distribution, thereby enhancing data privacy. Leveraging the post-processing property of differential privacy, as elucidated in \cite{dwork2014algorithmic}, ensures that any subsequent operation on a differentially private output remains privacy-preserving. Thus, data generation under the guidance of the pre-trained teacher model is deemed secure. Specifically, random Gaussian noise is incorporated during the teacher model's training as follows:
\begin{equation}
g_{T}\leftarrow g_{T}+N(0,\sigma _{n}^{2}c_{g}^{2}I)
\end{equation}
Here, $g_{T}$ represents the teacher's gradients, $\sigma _{n}$ is the noise scale, and $c_g$ signifies the gradient function's sensitivity. Subsequently, the teacher model's weight parameters are updated and truncated within the range (-c, c) to optimize the model:
\begin{equation}
w \leftarrow clip(w+\alpha \cdot Adam(w,g_T),-c,c)
\end{equation}
For practical implementation, we use Opacus \cite{yousefpour2021opacus}, Facebook's specialized library for training PyTorch models with differential privacy.

\begin{figure*}[t]
    \centering
    \includegraphics[width=\linewidth]{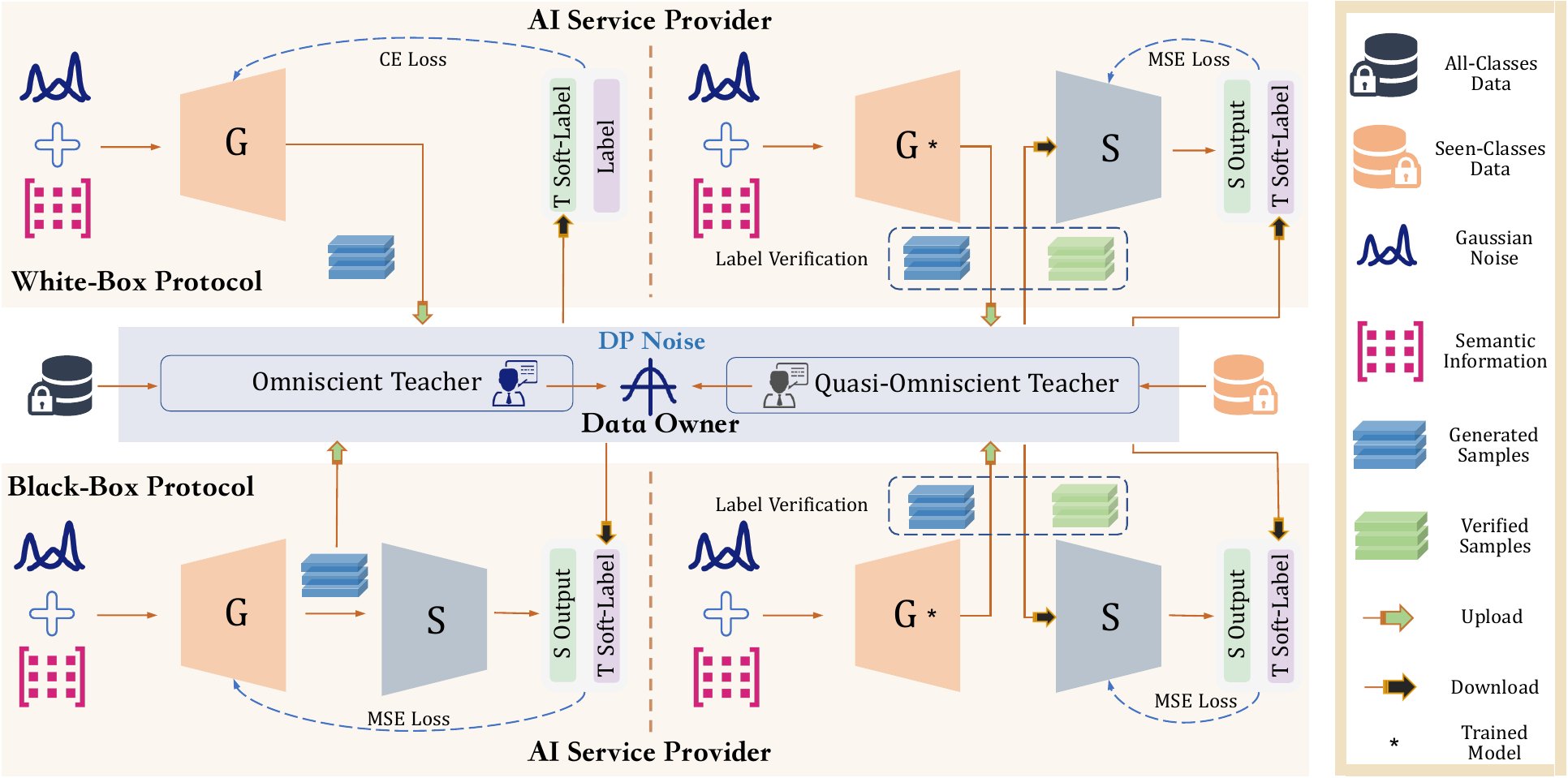}
    \caption{The overarching paradigm for both black-box and white-box protocols. In the white-box protocol, the generator accesses teacher weights during training, whereas in the black-box protocol, only output guidance from the teacher is utilized.}
    \label{AZSL-framework}
\end{figure*}
\subsection{\textbf{Dual Training Protocols}}

To address the multifaceted needs of data owners concerning both privacy preservation and performance optimization, two distinct training protocols have been devised, each characterized by a unique security level: the white-box and black-box protocols. Fig.\ref{AZSL-framework} provides a comprehensive visualization of the SG-ZSL paradigm within the context of both protocols. Each protocol encompasses three core components: 1) the isolated secure data and teacher model located at the data owner's domain; 2) the student model and generator positioned within the AI service provider's domain; and 3) the information exchange channels. For the white-box protocol, the teacher's weights are utilized in computing the gradient for both the generator and student network training. In contrast, the black-box protocol relies solely on the teacher model to furnish softmax output as pseudo labels, thereby excluding it from backpropagation during the SG-ZSL paradigm optimization.

\subsubsection{\textbf{White-Box Protocol}}

Under the white-box protocol, the data owner's pre-trained teacher model provides both gradient and softmax output to guide the training of models in the AI service providers, as explained below:

\noindent{\textbf{(i) Uploading generated data:}} Under the guidance of the teacher model, the generator produces class-specific features using noise vector $\mathbf{z}$ and class-level semantic embedding $\mathbf{a}$ (attributes or BERT model representations of class names \cite{devlin2018bert}). The synthesized features are represented as $\tilde{x} = G(z|a; \theta_{G})$, with the objective of generating superior synthetic data instead of real data to train the student model.

\noindent{\textbf{(ii) Gradient and softmax guidance:}} The teacher model, upon receiving $\tilde{x}$, processes it through the loss function:
\begin{equation}\label{5}
    \min_{\theta_{G}}  \mathcal L(\tilde{x},y;\theta_G)+ \alpha \mathcal R(\tilde{x}),
\end{equation}
where $\mathcal L(\cdot)$ signifies the cross-entropy loss for classification by the teacher model, and $\mathcal{R}(\cdot)$ denotes the regularization term during feature generation. This regularization, executed at the data owner's domain, aims to minimize the distribution discrepancy between real and generated features, ensuring the AI service provider remains oblivious to the real data.

\noindent{\textbf{(iii) Feedback downloading:}} A request is dispatched to the AI service provider to retrieve the gradient, the regularization of distribution divergence, and the softmax output.

\noindent{\textbf{(iv) Label verification:}} 
Using the softmax output, pseudo labels are computed, and misclassified samples are filtered using label verification as follows:
\begin{equation}\label{3}
\begin{aligned}
(\tilde{x}^{*},y^{*}) \in \{&(\tilde{x}, y)|y=\mathop{argmax}T(\tilde{x};\theta _{T}^*),\\
&\tilde{x}=G(z|a;\theta _{G}^*)\},
\end{aligned}
\end{equation}
where $T$ represents the teacher model, and $\tilde{x}^{*}$ and $y^{*}$ denote the filtered high-quality generated features and their corresponding class labels, respectively.

\noindent{\textbf{(v) Training the student model:}} Then the student model's training is articulated as follows:
\begin{equation}
\label{4}
\setlength{\abovedisplayskip}{5pt} 
\setlength{\belowdisplayskip}{2pt}
    \min_{\theta_{S}}  \left \| T^{*} (\tilde{x}^{*};\theta_{T}^{*})-S(\tilde{x}^{*};\theta_{S})\right \|_{2}^{2}
\end{equation}

In this protocol, the gradient is directly applied to the generated features, markedly augmenting the generator's efficacy. However, the gradient feedback is deemed mid-risk information, potentially revealing details of the teacher model, while the softmax output and regularization feedback are categorized as low-risk.

\subsubsection{\textbf{Black-Box Protocol}}

The black-box protocol, in contrast to its white-box counterpart, restricts the teacher model's guidance during the second step.  
Specifically, only the softmax output and regularization can be solicited from the teacher model, ensuring the teacher model's weights remain inaccessible and that it is not involved in the backpropagation optimization process, thereby reducing the risk of model leakage. The black-box protocol is elucidated step-by-step below:

\noindent{\textbf{(I) Uploading generated data:}} Analogous to the white-box protocol, the generated features are represented as $\tilde{x} = G(z|a; \theta_{G})$, which are subsequently transmitted to the data owner for processing.

\noindent{\textbf{(ii) Softmax guidance:}} Upon receipt of the generated features $\tilde{x}$, the data owner calculates the softmax output and divergence regularization. Thereafter, a request is initiated to relay the feedback to the AI service provider.

\noindent{\textbf{(iii) Label verification:}} 
The generated data is then evaluated to ensure the conditional class input matches the teacher's softmax output, with misaligned samples discarded.

\noindent{\textbf{(iv) End-to-end training:}} The generator $G$ and student network $S$ undergo end-to-end training as outlined in the objective function:
\begin{equation}\label{2}
    \min_{\theta_{G},\theta_{S}}  \left \| T^{*} (\tilde{x};\theta_{T}^{*})-S(\tilde{x};\theta_{S})\right \|_{2}^{2}+\alpha \mathcal{R}(\tilde{x}),
\end{equation}
The comprehensive training procedures for both protocols are delineated in Algorithm \ref{white-black}.

\begin{algorithm}[t]
    \small
    \caption{Training Procedure in Both Protocols}     
    \label{white-black}
    \begin{algorithmic}[1] 
    \REQUIRE ~~\\  
     Pre-trained Teacher network $\theta_{T}^*$, class labels $\mathcal Y_{tr}$ and their auxiliary semantic embedding $\mathcal{A}$; the maximal number of training epochs $T_g$ and $T_s$ for generator and student network, respectively.
    \ENSURE ~~\\  
    The learned parameters $\theta_{G} $, $ \theta_{S}$ for generator $G$ and student network $S$, respectively. 
    
    \STATE Initializing, $\theta _{G} $, $ \theta _{S}$. Set the iteration epoch $t_g=1,t_s=1$.
    
    \WHILE{ $t_g<T_g$ }  
    \IF{White-Box Protocol}
    \STATE Training generator with gradient guidance of teacher network through Eq.(\ref{5}).
    \ELSIF{Black-Box Protocol}
    \STATE Training generator with output guidance of teacher network through Eq.(\ref{2}).
    
    \ENDIF
    \STATE $t_g:=t_g+1$;
    \ENDWHILE  
    \STATE Conducting label verification through Eq.(\ref{3}).
    
    \WHILE{ $t_s<T_s$ }  
    \STATE Training student network with output guidance of teacher through Eq.(\ref{4}).
    \STATE $t_s:=t_s+1$;
    \ENDWHILE   
    \end{algorithmic}  
\end{algorithm}

\subsection{\textbf{Absolute Zero-Shot Classification}}

In the testing phase, the omniscient teacher, having been trained on both seen and unseen features at the Data Owner's end, facilitates the generator in synthesizing features for all classes. Consequently, the student network is equipped to predict class labels for test features. Given these test features, the predicted class labels are determined as:
\begin{equation}\label{10}
\setlength{\abovedisplayskip}{2pt} 
\setlength{\belowdisplayskip}{2pt}
y^*=\mathop{argmax}\limits_{y \in \mathcal Y}p(y|x,\theta_{S}^{*}), 
\end{equation}
where $\theta_{S}^{*}$ represents the optimized parameters of the student model.

For the quasi-omniscient teacher model, the challenge confronting the student model intensifies. This heightened challenge arises because, during the training phase, neither the data owner nor the AI service provider possesses information regarding the unseen classes. In the testing phase, an initial step involves synthesizing a data batch for these unseen classes via the generator, denoted as $\tilde{x}=G(z|a; \theta_{G}^*)$, with $z$ indicating noise and $a$ representing the semantic embedding of the unseen class. 
Utilizing this synthesized data, the classifier $C$ undergoes training in a supervised learning task with the generated features, as formalized in the following equation:
\begin{equation}\label{12}
\min_{\theta_{C}}  -\mathbb{E}\left[ \log( y|\tilde{x}; \theta_{C})) \right],
\end{equation}
the function calculates the softmax loss by comparing the predicted label probabilities from synthesized features $\tilde{x}$ against actual labels $y$ to minimize the negative log-likelihood of correct class predictions, optimizing classifier $C$ for accurate unseen class label prediction.

Subsequently, the prediction of class labels for test features is executed as follows:
\begin{equation}\label{13}
y^*=\mathop{argmax}\limits_{y \in \tilde{\mathcal Y}}p(y|x,\theta_{C}^{*}),
\end{equation}
where $\mathcal{\tilde{Y}} = \mathcal Y_u$ is designated for the conventional ZSL task, and $\mathcal{\tilde{Y}} = \mathcal Y_s \cup \mathcal Y_u$ for the GZSL task.

In the context of the first SG-ZSL work, this work primarily seeks to address the ensuing research questions:
\begin{itemize}
    \item \textbf{RQ1:} How does the variation in teacher feedback influence the quality and diversity of the synthesized data?
    \item \textbf{RQ2:} How does the alteration in semantic information, when employed as generative conditions, affect the student model's performance?
    \item \textbf{RQ3:} Compare with the traditional ZSL methods, how do the SG-ZSL perform under the black-box and white-box protocols in terms of data privacy, model security, and classification accuracy?
    \item \textbf{RQ4:} Is the student model capable of transcending the constraints of the quasi-omniscient teacher model to generate novel knowledge (on unseen class)?
    \item \textbf{RQ5:} Does the SG-ZSL paradigm, which trains on both seen and unseen classes using synthesized data, enhance the congruence between seen and unseen classifiers in the GZSL challenge? Specifically, is there an improvement over prior ZSL approaches that employed real seen data and synthesized unseen data, potentially introducing a bias towards seen classes?


\section{Experiments}

\end{itemize}
\begin{table*}[htbp]
\caption{Detailed dataset statistics and data split in SG-ZSL. Notation: `att' - attribute; `S' - seen class; `U' - unseen class; `Om' - omniscient teacher; `Q-Om' - quasi-omniscient teacher. }
\begin{center}
\label{Dataset-ps}
\small

\renewcommand\arraystretch{0.6}

\resizebox{0.9\textwidth}{!}
{
\begin{tabular}{m{1.8cm}<{\centering}m{2.2cm}<{\centering}m{2cm}<{\centering}m{1.2cm}<{\centering}|m{1.5cm}<{\centering}m{1.5cm}<{\centering}|m{1cm}<{\centering}m{1cm}<{\centering}|m{1cm}<{\centering}m{1.5cm}<{\centering}}
\toprule 

\textbf{Dataset}&   {\textbf{Semantics}} &\textbf{\textbf{Class Number}} &{\textbf{Image}}& \multicolumn{2}{c|}{\textbf{Teacher}}& \multicolumn{2}{c|}{\textbf{SG-ZSL Training}} & \multicolumn{2}{c}{\textbf{SG-ZSL Evaluation}} \\
&&& &\multicolumn{2}{c|}{\textbf{(Om/Q-Om)}}&&&\multicolumn{2}{c}{\textbf{(Om/Q-Om)}}\\[3pt]

&&\textbf{S/U}& &\textbf{S} &\textbf{U} &   \textbf{S} &   \textbf{U} & \textbf{S} &\textbf{U}\\
\midrule  

AWA1 \cite{lampert2013attribute}  &BERT/att & 40/10 & 30475 & 19832&4542/0 & \textbf{0} &\textbf{0} & 4958 & 1143/5685\\
AWA2 \cite{xian2017zero}  &BERT/att  & 40/10 & 37322& 23527 & 6328/0 & \textbf{0} & \textbf{0} & 5882 & 1585/7913\\
aPY \cite{farhadi2009describing}  &BERT/att  & 20/12 & 15539 & 5932 & 6333/0 &  \textbf{0} & \textbf{0} & 1483 & 1591/7924\\

\bottomrule 
\end{tabular}
}
\end{center}
\end{table*}

\subsection{\textbf{Datasets}}
Our SG-ZSL model is evaluated on three benchmark datasets: AWA1 \cite{lampert2013attribute}, AWA2 \cite{xian2017zero}, and aPY \cite{farhadi2009describing}. Both AWA1 and AWA2 encompass 30,475 and 37,322 images, respectively, distributed across 50 classes. The aPY dataset contains 15,539 images spanning 32 classes. For semantic representation, embeddings generated by the BERT language model \cite{devlin2018bert} are adopted, with a consistent dimensionality of 768 across all datasets. The data splits differ based on the type of teacher model. For quasi-omniscient teachers, we adopt the data split proposed in \cite{xian2017zero}, wherein only seen class data is accessible to the teacher. Conversely, the omniscient teacher is trained across all classes. In alignment with prior ZSL studies \cite{zhang2020deep}, unseen classes are randomly divided into training and test sets. Comprehensive dataset details and SG-ZSL data splits are presented in Table \ref{Dataset-ps}.

\subsection{\textbf{Implementation Details}}
For image representation, 2048-dimensional ResNet101 features \cite{he2016deep} are utilized, consistent with \cite{xian2017zero}. Within our proposed paradigm, all networks are constructed using Multi-Layer Perceptrons equipped with LeakyReLU activations \cite{Xu2015Empirical}. Both the teacher and student models share the same architecture comprising two hidden layers with 1024 and 512 units, respectively. The generator contains a single hidden layer with 4096 hidden units and its output layer is ReLU. During the training process, we adopt the Adam optimizer and the learning rate of each network is set to $10^{-5}$. The dimension of the noise vector $z$ is a hyper-parameter, which is empirically set to 20 for all datasets. The weight of the regularization term is empirically set to 0.5 for AWA1 and AWA2, and 1 for aPY. A trade-off between accuracy and computational efficiency is taken into consideration when determining the number of generated features. In practice, we generate 400 synthetic features on average per class for all datasets. 
For the training epochs $T_{g}$ and $T_{s}$, we selected values that balance convergence and prevent overfitting or underfitting for both the generator and student network. Experimentally, we found performance plateaus in both networks beyond certain iterations, indicating an optimal stopping point for training. Consequently, $T_{g}=50$ and $T_{s}=80$ were set to optimize both computational efficiency and model effectiveness.
\subsection{\textbf{Evaluation Protocol}}
We follow the evaluation metrics proposed in \cite{xian2017zero}. For conventional ZSL tasks, we use the per-class average top-1 accuracy to evaluate classification performance to alleviate the data imbalance of classes. For the GZSL task, we use harmonic mean $H=(2 \times u \times s)/(u+s)$ for evaluation, where $u$ and $s$ denote average per-class top-1 accuracy on unseen and seen classes, respectively. It is noteworthy that existing methods aim to classify unseen data into corresponding unseen classes in conventional ZSL tasks, while the class space at test time involves both unseen and seen classes in SG-ZSL with the omniscient teacher. This makes SG-ZSL with an omniscient teacher more difficult compared with existing ZSL methods.

\subsection{\textbf{Main Results}}

\begin{table*}[t]
  \caption{Comparison results with the state-of-the-art methods in CZSL and GZSL tasks. CZSL measures per-class average top-1 accuracy (T1) on unseen classes. GZSL measures u = T1 on unseen classes, s = T1 on seen classes, H = harmonic mean. `WB’ \& `BB’: white- \& black-box protocol; `Om’ - omniscient teacher, `Q-Om’ - quasi-omniscient teacher. `SG-ZSL+WB/BB*' and `SG-ZSL+WB/BB' represent our model with omniscient and quasi-omniscient teachers, respectively. The best results are in bold.}\label{AZSL-ALL}
\begin{center}    
\renewcommand\arraystretch{0.5}
\tiny
\resizebox{\textwidth}{!}
{

\begin{tabular}{ccccc|ccc|ccc|ccc}
\toprule
&\textbf{Method} & \multicolumn{3}{c|}{\textbf{Zero-Shot Learning}} & \multicolumn{9}{c}{\textbf{Generalized Zero-Shot Learning}}\\

 &&  \textbf{AWA1}& \textbf{AWA2}  & \textbf{aPY}  &\multicolumn{3}{c}{\textbf{AWA1}} &\multicolumn{3}{c}{\textbf{AWA2}} &\multicolumn{3}{c}{\textbf{aPY}}\\
&&T1&T1&T1& u & s &H &u & s &H & u & s &H \\

\midrule 
\multirow{11}{*}{\textbf{}}  
&IAP \cite{lampert2013attribute} &35.9&35.9&36.6 &2.1 &78.2 &4.1 &0.9 &87.6 &1.8 & 5.7 &65.6 &10.4\\ 
&DAP \cite{lampert2013attribute}  &  44.1& 46.1 & 33.8 &0.0 & 88.7 &0.0 &0.0 &84.7& 0.0  &4.8 &78.3&9.0\\
&ALE \cite{akata2013label}  &    59.9&    62.5    & 39.7 &16.8  &76.1  &27.5  &14.0  &81.8  &23.9  &4.6  &73.7  &8.7\\
&DEVISE \cite{frome2013devise}   &54.2&    59.7&  39.8&13.4     &68.7 &22.4 &17.1 &74.7 &27.8 &4.9 &76.9 &9.2\\
&CONSE \cite{norouzi2013zero}  &45.6    &44.5  &  26.9&0.4 &88.6 &0.8 &0.5 & 90.6 &1.0  &0.0 &\textbf{ 91.2} &0.0\\
&ESZSL \cite{romera2015embarrassingly}&58.2    &58.6&    38.3  &6.6 &75.6 &12.1 &5.9 &77.8 &11.0  &2.4& 70.1 &4.6\\
&SYNC \cite{changpinyo2016synthesized} &54.0    &46.6&    23.9&8.9 &87.3 &16.2 &10.0 &90.5 &18.0  &7.4 &66.3 &13.3\\
&DEM \cite{zhang2017learning}  & 68.4 & 67.1& 35.0   & 32.8 & 84.7 & 47.3 & 30.5&  86.4 & 45.1  & 11.1 & 75.1 & 19.4 \\
&f-CLSWGAN \cite{xian2018feature}  & 68.2 & -  & -& 57.9 & 61.4 & 59.6 & - &-&- & -&-&-\\
&CE-GZSL \cite{han2021contrastive} & 71.0 &70.4&-& 65.3 &73.4 &69.1 &63.1 &78.6 &70.0&-&-&-\\
&SDGZSL \cite{chen2021semantics} &-&74.3&47.0 &-&-&-&69.6&78.2 &73.7&39.1&60.7 &47.5\\
&ICCE \cite{kong2022compactness} &74.2 &72.7&49.5 & 67.4 &81.2 &73.6 & 65.3 &82.3& 72.8 &45.2&46.3 &45.7  \\
\midrule
\multirow{7}{*}{\textbf{}}
&DTN \cite{zhang2020deep} & 69.0 &-& 41.5 & 54.8 & 88.5 &67.7 &-&-&- &37.4 &87.9 &52.5\\ 
&GMSADE \cite{gune2020generative} &81.3 & 80.7 &49.9 & 71.2  &87.7& 78.6 &71.3&86.1&  78.0 &76.1 &39.3 &51.8\\
&EDE \cite{zhang2020towards} & \textbf{85.3} &77.5& 31.3 & 71.4 &\textbf{90.1}& 79.7 &68.4 &\textbf{93.2 }&78.9&29.8 &79.4 &43.3\\

&BGT \cite{li2021bidirectional}&-&\textbf{82.4}&49.8  &-&-&- &56.2 &82.2 &66.7&39.3 &72.9&51.0  \\

\midrule 

&Q-Om Teacher& 0.0 & 0.0 & 0.0 & 0.0 & 92.9 & 0.0 & 0.0 & 93.1 & 0.0 & 0.0 & 91.6 & 0.0 \\ 
&Om Teacher & 92.1 & 91.7 & 90.8 & 92.1 & 92.5 & 92.3 & 91.7 & 92.2 & 91.9 & 90.8 & 91.4 & 91.1\\
\midrule 
&\textbf{SG-ZSL+BB} &14.1 & 19.9 & 12.3 &   4.1 & 3.7& 3.9 & 3.5&3.7 &3.6& 6.8 &4.0 & 5.1\\
&\textbf{SG-ZSL+WB} & 34.5 & 36.5 & 18.7 & 23.4 &34.3 &27.8 & 27.3 &44.3 &33.7 &17.9 & 52.5 &26.7 \\
&\textbf{SG-ZSL+BB*} & 33.5 & 29.0 &30.2 &33.5 & 28.6 & 30.9 &29.0 & 25.3& 27.0& 30.2&42.2 & 35.2 \\
&\textbf{SG-ZSL+WB*} &77.9 & 79.0& \textbf{83.9} & \textbf{77.9} &81.8&\textbf{79.8 }& \textbf{79.0}&86.7&\textbf{82.7} & \textbf{83.9}& 85.7 &\textbf{84.8} \\

\bottomrule
\end{tabular}
}
\end{center}
\end{table*}

\subsubsection{\textbf{Comparisons with State-of-Arts}}
Table \ref{AZSL-ALL} presents results for both CZSL and GZSL tasks. Given that this is the inaugural SG-ZSL study, a comparison with traditional state-of-the-art methods serves as a reference.
The selected methods can be categorized into inductive and transductive ZSL methods. Methods in the upper part of Table \ref{AZSL-ALL}, \ie, IAP, are inductive ZSL methods, which access only labeled seen class data during the training process. The rest of the four methods, \ie, DTN, are transductive methods, which utilize both labeled seen class data and unlabeled unseen class data for model training.

To investigate \textbf{RQ1}, we show results under two kinds of feedback from omniscient and quasi-omniscient teachers.
SG-ZSL student model with omniscient teacher achieves promising performance in both CZSL and GZSL in the white-box protocol. 
We achieve the best performance in GZSL, especially on aPY, with an increase in harmonic mean of 32.3\%, which indicates an improved balance of seen and unseen classes. 
As for the black-box protocol, the accuracy on unseen classes is 4.9\% higher than on seen classes on AWA1. It indicates that the SG-ZSL student model is promising to mitigate the class-level overfitting issue in the GZSL task proposed in \textbf{RQ5}.
Compared with inductive ZSL methods, results show that our model with the quasi-omniscient teacher in a white-box protocol gains satisfactory performance in GZSL, especially on aPY, with 7.3\% higher performance on the harmonic mean compared with non-generative inductive ZSL methods. 
Despite the quasi-omniscient teacher model's inability to recognize unseen classes and the student model's lack of access to real seen and unseen data, our student model still secures robust accuracy across various ZSL scenarios. For example, it achieves 34.5\% accuracy in inductive ZSL settings on AWA1 and a harmonic mean of 26.7\% in GZSL on the aPY dataset.  This underscores the student model's capacity to extrapolate and generalize from the teacher's knowledge without data exposure, as explored in \textbf{RQ4}. Additionally, when contrasted with traditional TZSL methods, our model exhibits significant accuracy enhancements in GZSL, especially for unseen classes (\ie demonstrate a 6.5\% and 7.8\% improvement on AWA1 and aPY datasets, respectively), and presents a reduced discrepancy between seen and unseen class accuracies, showcasing an advanced ability to mitigate seen class bias as mentioned in RQ5.
For the black-box protocol, results show our SG-ZSL student model outperforms random guessing, which is around 10\% on AWA1, AWA2, and 8\% on aPY. The white-box protocol demonstrates better performance than the black-box protocol for the student, indicating that gradient guidance provides more information.

\begin{table*}[h]
\begin{center}    
\caption{Experimental results in black-box protocol with the omniscient teacher in both CZSL and GZSL tasks.}
\label{AZSL-all-blackbox}
\renewcommand\arraystretch{0.5}
\tiny
\resizebox{\textwidth}{!}
{
\begin{tabular}{cccc|ccc|ccc|ccc}
\toprule  
\textbf{Method} & \multicolumn{3}{c|}{\textbf{Zero-Shot Learning}} & \multicolumn{9}{c}{\textbf{Generalized Zero-Shot Learning}}\\

 &  \textbf{AWA1}& \textbf{AWA2}  & \textbf{aPY}  &\multicolumn{3}{c}{\textbf{AWA1}} &\multicolumn{3}{c}{\textbf{AWA2}} &\multicolumn{3}{c}{\textbf{aPY}}\\
&T1&T1&T1& u & s &H &u & s &H & u & s &H \\

\midrule 

Label-Conditioned & 15.5&10.0 & 7.0&15.5&24.3& 18.9&10.0&17.8&12.8&7.0& 3.8&4.9\\
Attribute-Conditioned & 10.1 &23.0& 8.2 &10.1& 11.3&10.7&23.0&17.6&20.0&8.2& 5.0 & 6.3\\
w/o Label Verification & 25.6 & 24.7 & 11.8 & 25.6 &15.6& 19.4&24.7&18.1&20.9&11.8&20.9&15.0\\
w/o Regularization &26.8 & 23.7&23.2 & 26.8 &26.7&26.8&23.7&23.2 & 23.4& 23.2&25.6&24.3\\
\midrule  
\textbf{SG-ZSL+BB} & \textbf{33.5} & \textbf{29.0 }&\textbf{30.2 }&\textbf{ 33.5} & \textbf{28.6} & \textbf{30.9 }&\textbf{29.0} & \textbf{25.3 }&\textbf{27.0 }& \textbf{30.2}&\textbf{ 42.2} & \textbf{35.2} \\

\bottomrule 
\end{tabular}}
\end{center}
\end{table*}
\subsubsection{\textbf{Comparisons in Black-Box Protocol}}
 
As it is the first time to propose this setting, we provide several baselines for comparison in Table \ref{AZSL-all-blackbox}. We provide labels and attributes for conditional feature generation to investigate \textbf{RQ2}. 
Our proposed paradigm with BERT embedding achieves the best performance, \ie, with 18.0\% and 23.4\% increases in unseen accuracies on AWA1 compare with label-conditioned and attribute-contribution separately. Results show that our paradigm gains noticeable improvement in accuracy with label verification, \ie, with 20.2\% higher performance on harmonic mean on aPY dataset. And results indicate the effectiveness of adopting regularization, \ie, it achieves 3.6\% and 10.9\% increases in Harmonic mean on AWA2 and aPY. The comparison with baselines demonstrates the effectiveness of our SG-ZSL student model in a black-box protocol with the omniscient teacher.

\subsubsection{\textbf{Performance vs Paradigm Privacy}}
Compared to traditional ZSL methods, the performance under the white-box protocol is very promising, since data privacy is already preserved and our model can still achieve adequate performance. Compare with the white-box protocol, 
the black-box protocol indeed operates under a more constrained information flow, where only softmax outputs from the teacher model are used as pseudo-labels for the student model, without direct gradient exchange. This design choice inherently poses challenges to optimization efficiency compared to direct gradient-based methods. However, this constraint is a deliberate design choice to enhance privacy.
Thus, the performance of the black-box protocol is reasonable because both data privacy and model safety are guaranteed as proposed in \textbf{RQ3}. 

As for model copyright reservation, traditional ZSL methods often involve sharing model details across entities, raising potential issues related to intellectual property and copyright infringement. Our SG-ZSL paradigm circumvents these issues by utilizing a sentinel mechanism that facilitates the learning process without exposing the internal architecture of the models involved. This is achieved by guiding the generation of synthetic data as a medium for communication between the AI Service provider and the Data Owner, enabling both parties without directly sharing the models themselves. This approach ensures that copyright and intellectual property rights are respected and protected, offering a sustainable model for collaborative AI development and usage.

\subsection{\textbf{Analysis and Discussion}}
\subsubsection{\textbf{Feature Generation Regularization Analysis}}

\begin{table}[t]
\centering
  
\caption{Experimental results with different constraints for feature generation in GZSL task in the \textbf{white-box} protocol. ‘CE’ represents cross-entropy loss, `MMD' represents MMD distance loss, and `KL' represents KL divergence loss.}
\label{AZSL-loss}
\small
\renewcommand\arraystretch{0.9}

\begin{tabular}{cccc|ccc}
\toprule  
\textbf{Method} 
 &\multicolumn{3}{c}{\textbf{AWA2}} &\multicolumn{3}{c}{\textbf{aPY}}\\
  &u & s &H  & u & s &H \\

\midrule  

CE&76.1&83.8&79.8 &83.0&84.5&83.7\\
CE+MMD& \textbf{79.9} & 85.1 & 82.5 & 81.5 & 85.5 & 83.5  \\
CE+KL & 79.0&\textbf{86.7}&\textbf{82.7} & \textbf{83.9}& \textbf{85.7 }&\textbf{84.8}\\

\bottomrule 
\end{tabular}
\end{table}

\begin{table}[t]
\begin{center}    
\caption{Experimental results with different constraints for feature generation in GZSL task in the \textbf{black-box} protocol. ‘CE’ represents cross-entropy loss, `MMD' represents MMD distance loss, and `KL' represents KL divergence loss.}
\label{black-loss}
\renewcommand\arraystretch{0.9}
\resizebox{\linewidth}{!}
{

\begin{tabular}{c|ccc|ccc|ccc}
\toprule  
\textbf{Method 
} &\multicolumn{3}{c}{AWA1\textbf{}}&\multicolumn{3}{c}{\textbf{AWA2}} &\multicolumn{3}{c}{\textbf{aPY}}\\
  &u & s &H &u & s &H  & u & s &H \\

\midrule  

CE& 26.8 &26.7 &26.8& 23.7& 23.2& 23.4& 23.2& 25.6& 24.3\\
CE+MMD& 31.8 &25.3&28.2& \textbf{33.8} &20.5&25.5 & 26.0& 36.3& 30.3 \\
CE+KL & \textbf{33.5}& \textbf{28.6}& \textbf{30.9}& 29.0& \textbf{25.3}& \textbf{27.0}& \textbf{30.2}& \textbf{42.2}& \textbf{35.2}\\

\bottomrule 
\end{tabular}}

\end{center}
\end{table}
The key issue in our data-free knowledge transfer framework is to generate high-quality features, which are expected to have a similar distribution to real data. To show the influence of different constraints during the feature generation process, we provide analysis with different regularization terms for generator training in Table \ref{AZSL-loss}. KL and MMD loss \cite{gretton2012kernel} aim to minimize the distribution difference between real and generated features. Results show that adding distribution constraints of synthesized data is beneficial for feature generation. For example, the harmonic mean increases 2.7\% and 2.9\% with MMD and KL loss respectively compared with the baseline that only contains cross-entropy loss. Besides, results indicate that KL and MMD loss are both effective and KL loss performs better to a small extent, which shows the effectiveness of KL regularization.

We also provide an extensive analysis of the impact of different feature generation regularizations in the black-box scenario in Table \ref{black-loss}. Similarly, we provide MMD and KL loss as regularization for feature synthesis in the GZSL task as the regularization term is essential for the generalization ability of the SG-ZSL model. The experimental results show that the SG-ZSL model with regularization term outperforms the one with only cross-entropy loss, \ie, with 6\% and 10.9\% improvement on harmonic mean with MMD and KL loss on aPY, indicating the effectiveness of the constraint for feature generation. Besides, the SG-ZSL model with KL constraint achieves the best performance in harmonic mean, with 4.9\% and 2.7\% increases on aPY and AWA1 datasets respectively, which indicates that the SG-ZSL model with KL loss can make a better balance between seen and unseen classes. 

\subsubsection{\textbf{Student vs Teacher Performance Analysis}}
\begin{figure}[t]
    \centering
    \subfloat[AWA1]{
        \includegraphics[width=0.22\textwidth]{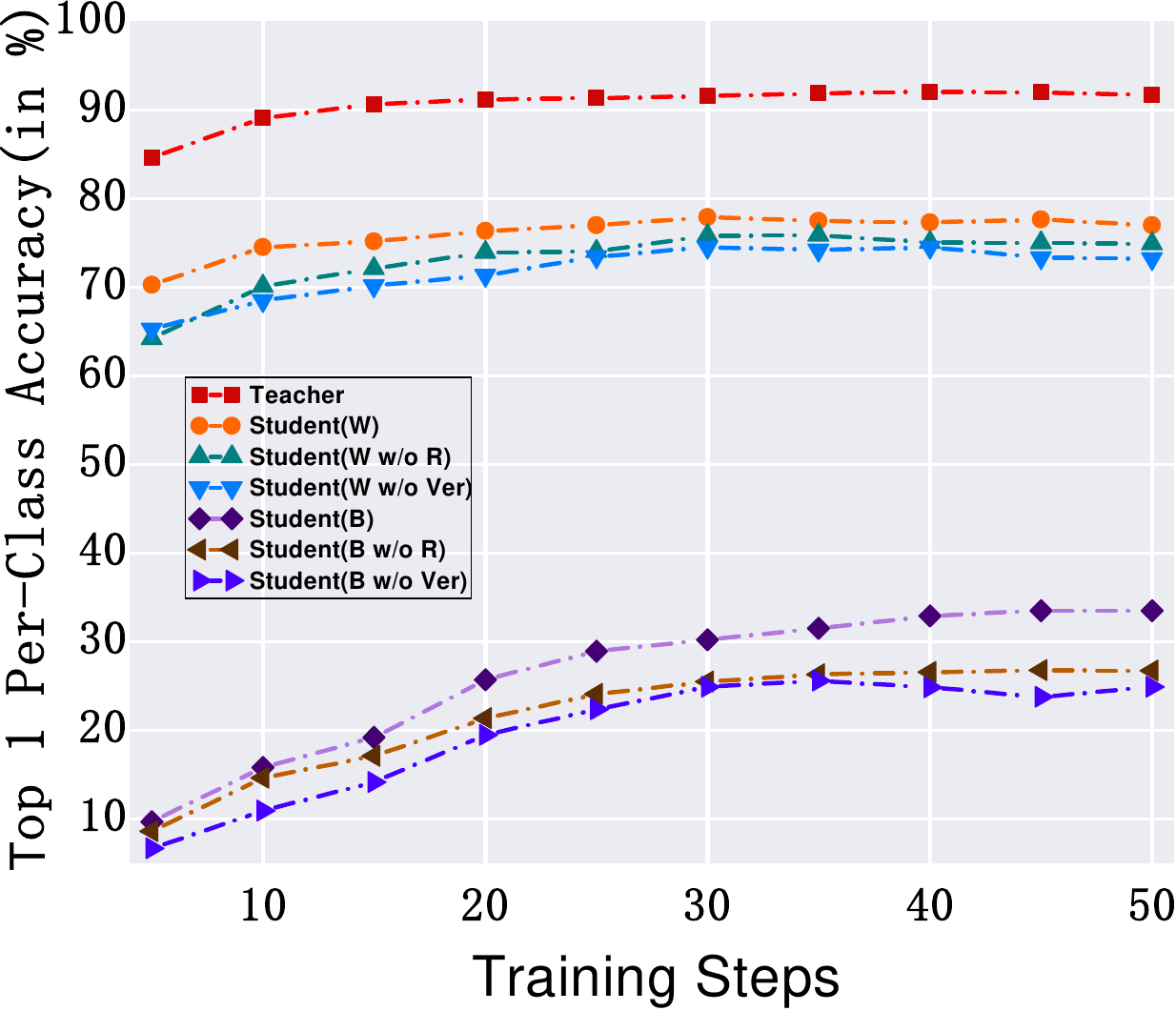}
     }
    \subfloat[aPY]{
        \includegraphics[width=0.22\textwidth]{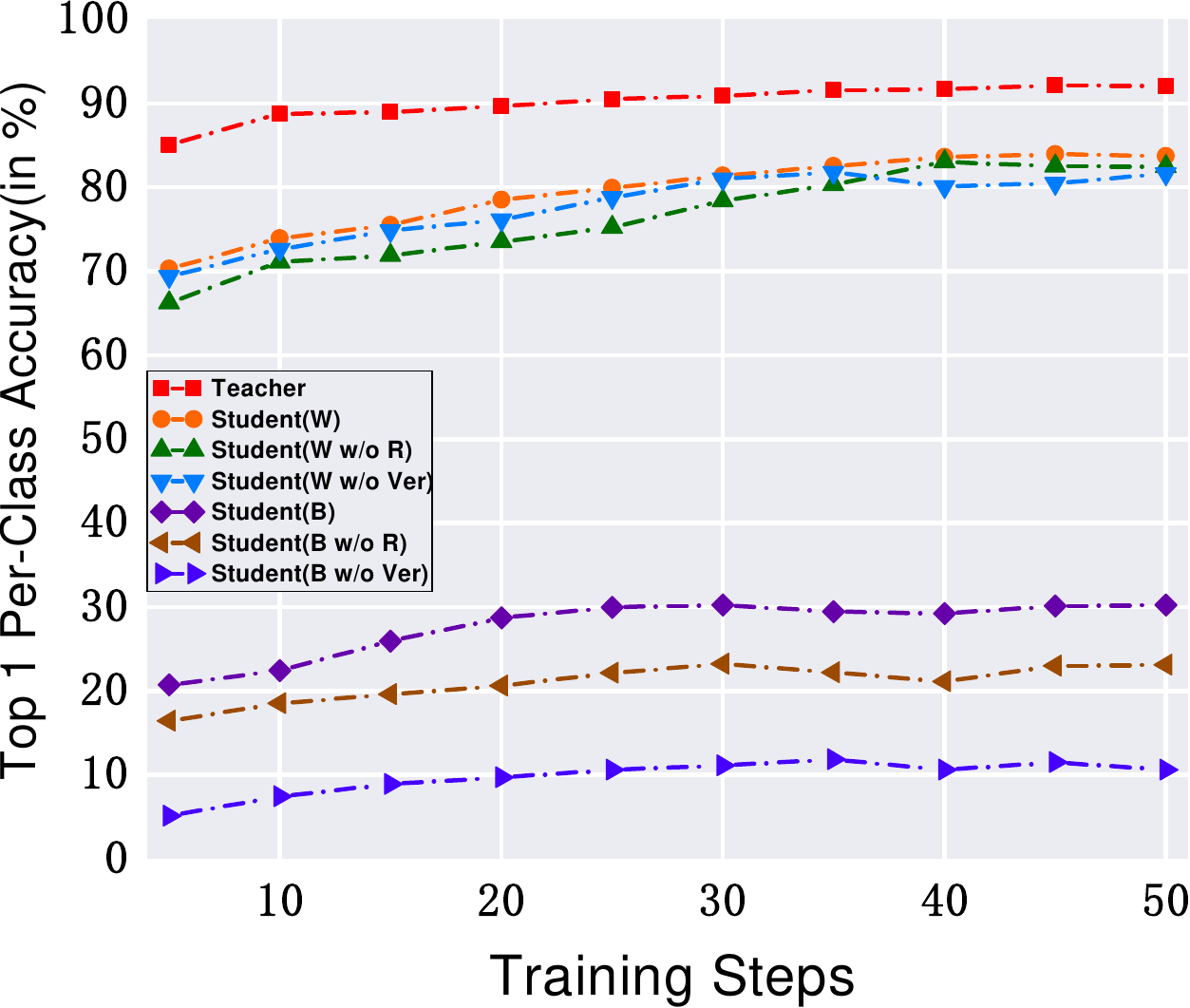}
    }
    \caption{Epoch analysis for unseen accuracy. `Ver': label verification. `R': regularization term.}
    \label{acc-t-s}
\end{figure}

Here, we undertake experiments to assess the relationship between teacher performance and student performance. The outcomes observed from the omniscient teacher and student models, across escalating training steps in both protocols on AWA1 and aPY, are depicted in Fig. \ref{acc-t-s}.

Operating under supervised learning with access to real data, our teacher model reliably guides student models, as evidenced by its performances of 92.1\% on the AWA1 dataset and 90.8\% on the aPY dataset. Within the white-box protocol, the student model approaches the performance of the teacher model, signifying the efficacy of the gradient guidance mechanism. Furthermore, our findings elucidate that the incorporation of regularization terms enhances the model's performance, showcasing the pivotal role of feature distribution throughout the training phase. Further, our analysis reveals that label verification in both protocols enhances performance, highlighting its necessity. 
This is attributed to its capacity to mitigate the negative effects resulting from the creation of low-quality features.

\begin{table}[t]
    \caption{Results in the white-box protocol with an omniscient teacher under different privacy budgets $\epsilon$. }\label{DP}
\begin{center}    
\renewcommand\arraystretch{0.5}
\tiny
\resizebox{0.45\textwidth}{!}
{
\begin{tabular}{ccccc}
\toprule  
 \textbf{Dataset}&\textbf{Accuracy} &\textbf{$\epsilon=30$}&\textbf{$\epsilon=50$}&\textbf{$\epsilon=\infty$} \\
\midrule

\multirow{2}{*}{\textbf{AWA1}}  
&Teacher Model & 56.7 & 68.4 & 92.1 \\
& Harmonic Mean & 41.7& 56.4&79.8\\
\midrule
\multirow{2}{*}{\textbf{AWA2}}  
&Teacher Model &59.1&70.5&91.7\\
& Harmonic Mean& 46.8& 60.3& 82.7\\
\midrule
\multirow{2}{*}{\textbf{aPY}}  
&Teacher Model &60.6&72.4&90.8\\
& Harmonic Mean & 43.6& 62.2&84.8 \\

\bottomrule 
\end{tabular}
}
\end{center}
\end{table}

\subsubsection{\textbf{Teacher Model Privacy Evaluation}}
Table \ref{DP} displays the performance corresponding to various privacy budgets $\epsilon$ when DP is incorporated into teacher training. Here, $\epsilon=\infty$ signifies the baseline non-private performance, i.e., absent DP in teacher training. The results demonstrate that larger $\epsilon$ values correspond to enhanced performances for both the teacher and student models, indicating that a smaller $\epsilon$ yields heightened data security protection. A trade-off between performance and privacy level is observed, allowing for an adjustment of the privacy budget to achieve a balance.

\begin{figure}[htbp]
    \centering
    \subfloat[AWA1*]{
        \includegraphics[width=0.47\linewidth]{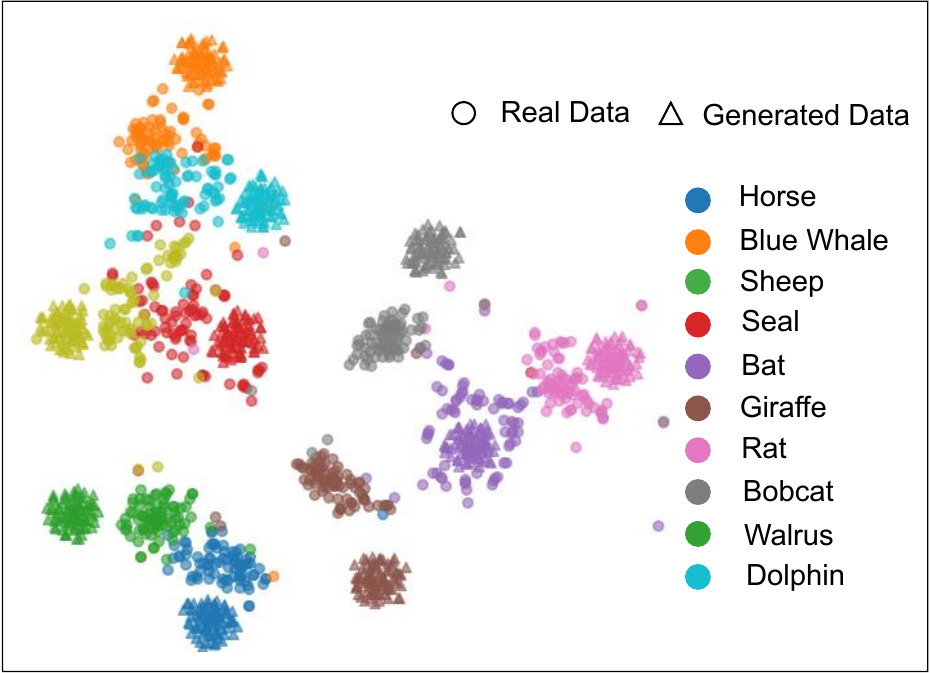}
     }
    \subfloat[aPY*]{
        \includegraphics[width=0.47\linewidth]{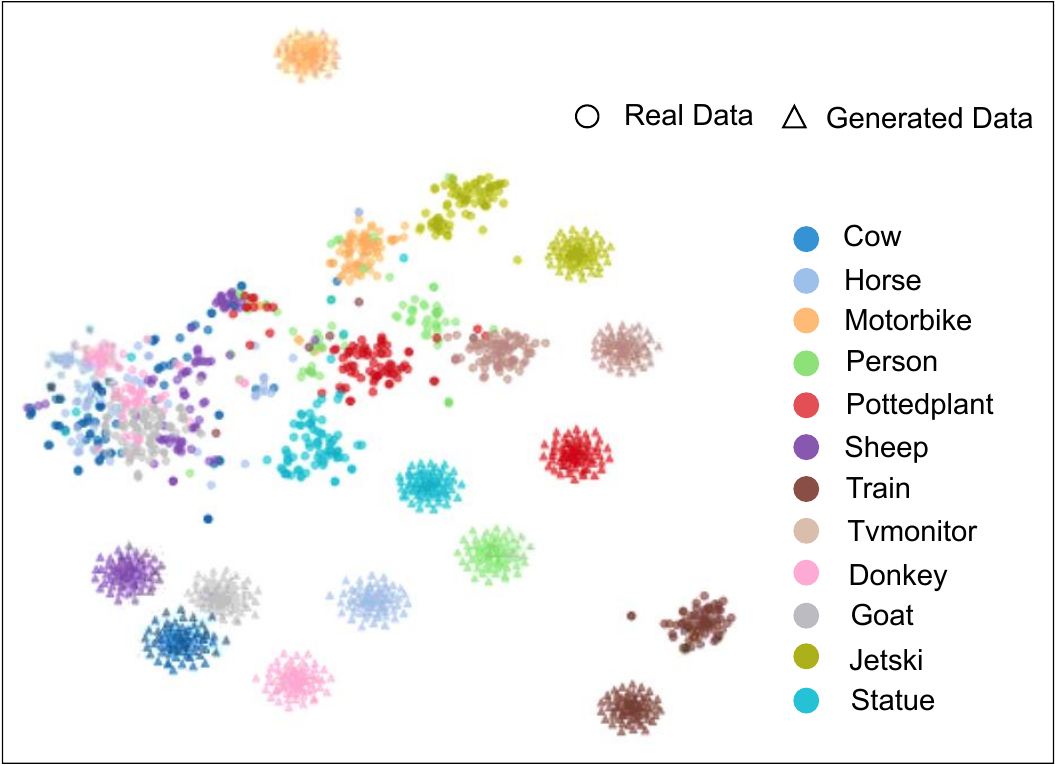}
    }
    \\
    \subfloat[AWA1]{
        \includegraphics[width=0.47\linewidth]{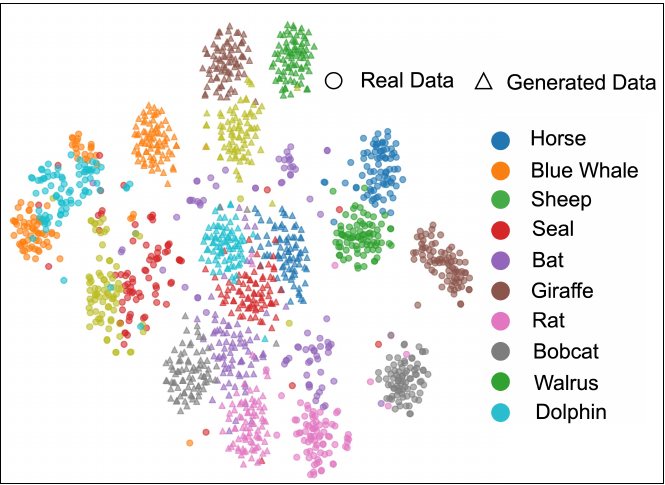}
     }
    \subfloat[aPY]{
        \includegraphics[width=0.47\linewidth]{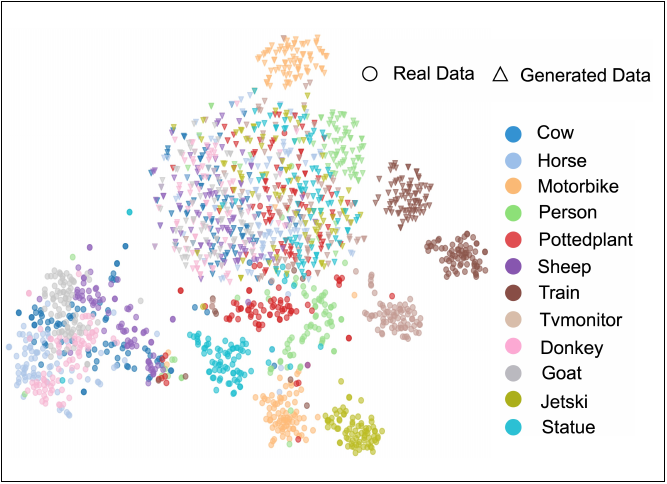}
    }

    \caption{The t-SNE visualization on AWA1 and aPY. All experiments are simulated under white-box protocol, with the synthetic features in (a) and (b) generated from generators that follow the omniscient teacher (indicated with *), and those in (c) and (d) generated from generators that follow the quasi-omniscient teacher.}
    \label{tsne}
\end{figure}
\subsubsection{\textbf{Quality of Generated Features}}
Fig. \ref{tsne} displays t-SNE visualizations of real and synthetic unseen features under the white-box protocol guided by two distinct teacher models across AWA1 and aPY datasets. For clarity in visualization, a subset of features is randomly selected. Features synthesized under the guidance of the omniscient teacher, illustrated in (a) and (b), closely emulate real feature distributions and demonstrate significant class clustering. This effectiveness, even without real data, highlights our model's capacity to generate class-coherent features in ZSL scenarios. 
In comparison, unseen class features synthesized under the guidance of the quasi-omniscient teacher, shown in (c) and (d), exhibit a slight decline in quality, \ie, the distribution of generated features is farther from the real data distribution, which illustrates the limitation of the generator. However, the model is trained without access to unseen class information. The capability of synthesizing unseen class data shows the potential to generate novel knowledge.

\begin{figure}[t]
    \centering
    \subfloat[AWA1]{
        \includegraphics[width=0.45\linewidth]{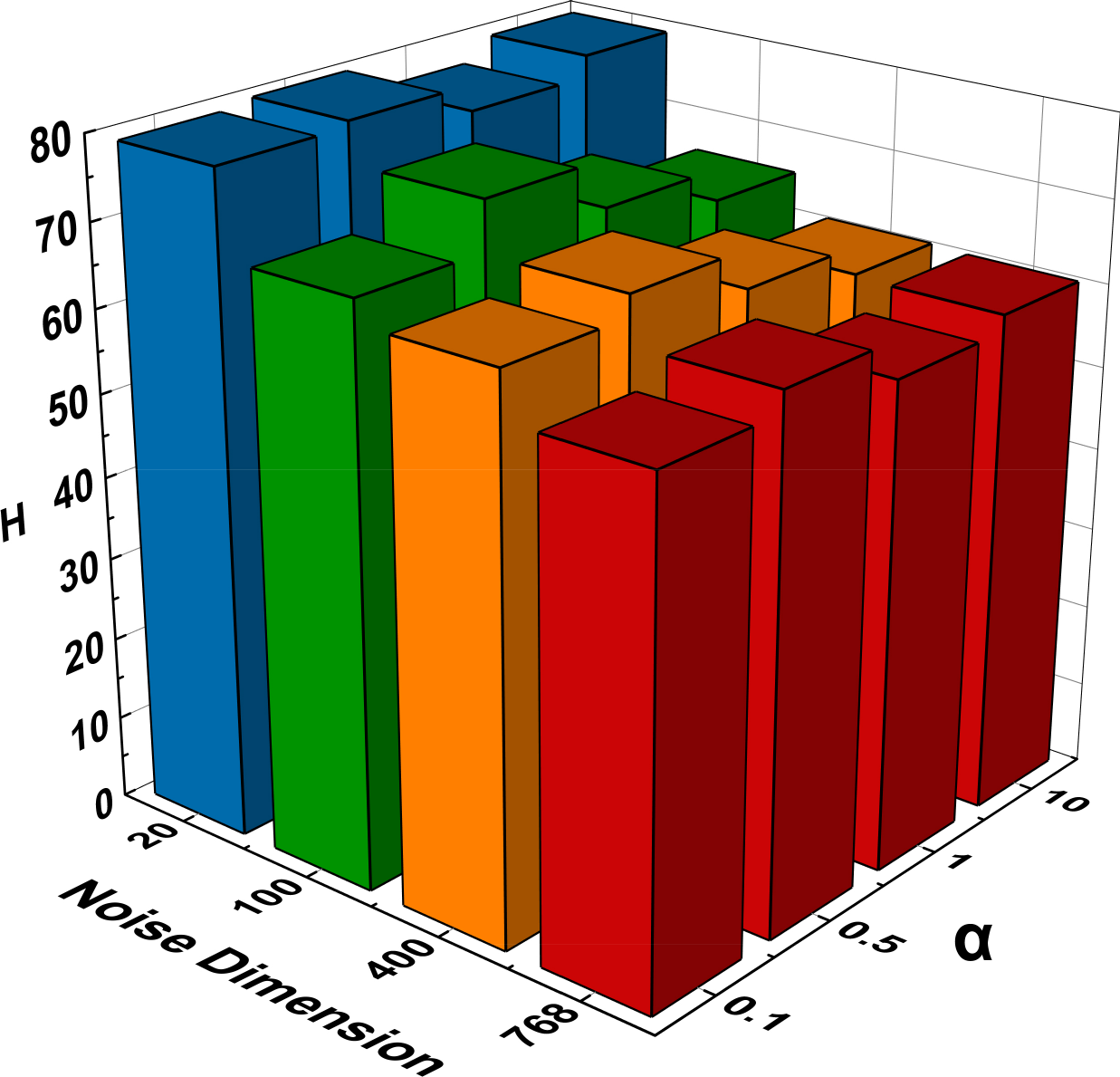}
     }
    \subfloat[aPY]{
        \includegraphics[width=0.45\linewidth]{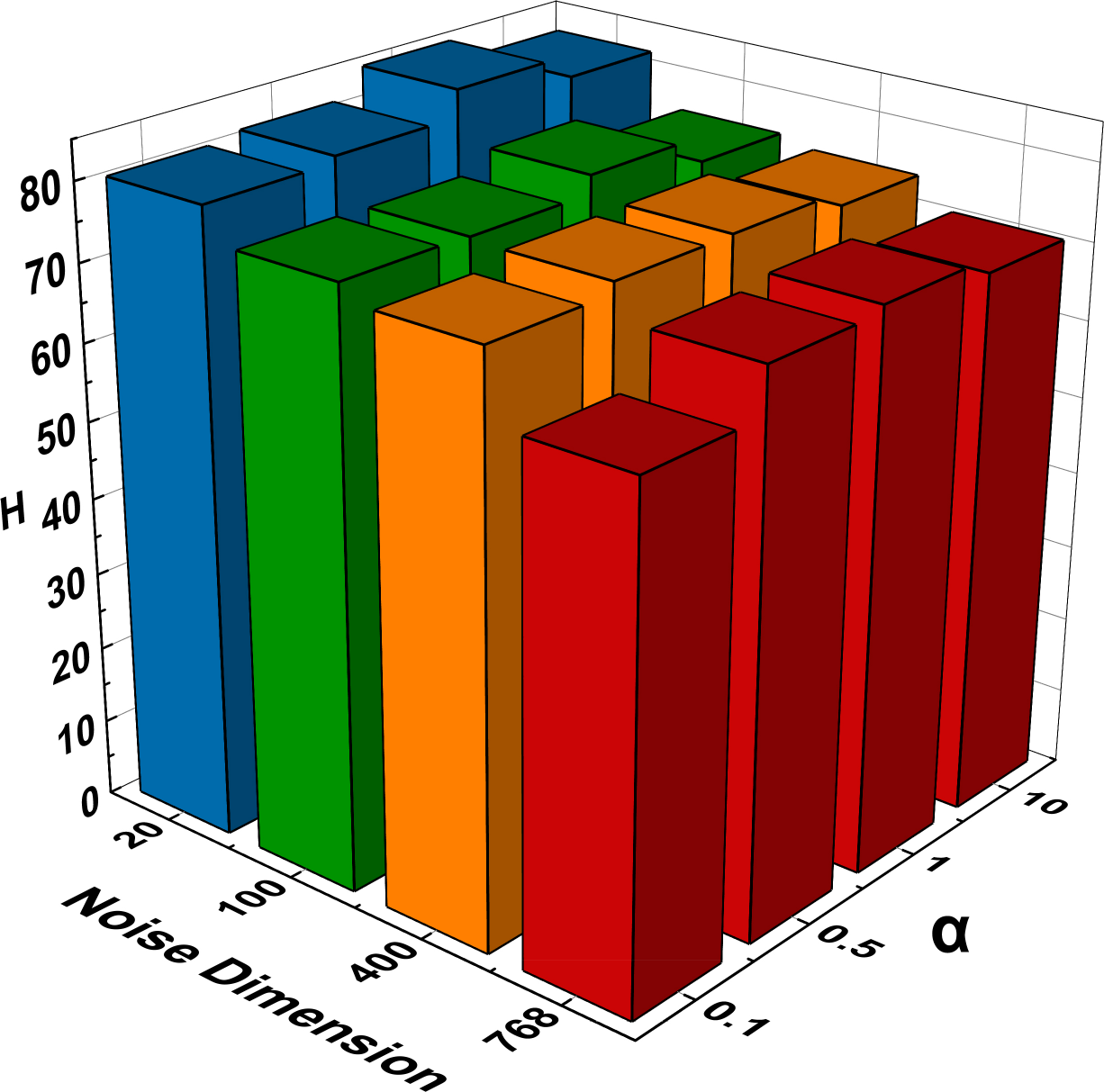}
    }
    \caption {Noise dimension and parameter $\alpha$ analysis with omniscient teacher in white-box protocol.}\label{hyper} 
\end{figure}

\subsubsection{\textbf{Hyper-Parameter Analysis}}
We assess the impact of two pivotal hyper-parameters, namely, noise dimension and regularization weight, on our student model. Two ablation studies are conducted on the AWA1 and aPY datasets within a white-box protocol framework, engaging an omniscient teacher, as illustrated in Fig. \ref{hyper}. We select four disparate noise dimensions 20, 100, 400, and 768 to elucidate their relationship with the harmonic mean. The findings reveal a performance decrement correlating with the expansion of the noise dimension across both datasets, suggesting that higher-dimensional noise may engender significant interference.
Concerning the regularization weight, we designate the values of $\alpha$ as 0.1, 0.5, 1, and 10 for the experimental analysis. As shown in Fig. \ref{hyper}, the harmonic mean on both datasets exhibits marginal fluctuation with varying $\alpha$ values. Optimal performance is attained at $\alpha$ values of 0.5 and 1 for AWA1 and aPY datasets, respectively, demonstrating a nuanced interaction between regularization weight and model performance.

\subsubsection{\textbf{Impact of Semantic Information}}
We further investigate the influence of various semantic embeddings on the GZSL task. The experimental analysis encompasses three distinct semantic typologies, namely, attributes, Word2Vec, and BERT, serving as the evaluation benchmarks. As delineated in Table \ref{WB-semantic}, the comparative outcomes across all three semantic modalities in the GZSL task are relatively aligned, manifesting the robustness of our model with respect to semantic embedding. Notably, the BERT embedding outperforms, signifying the superior efficacy of BERT representation in capturing semantic nuances.
\begin{table}[t]
\begin{center}    
\caption{Experimental results in white-box protocol with omniscient teacher using different semantic information in GZSL task.}
\label{WB-semantic}
\renewcommand\arraystretch{0.8}
\scriptsize
\resizebox{0.45\textwidth}{!}
{

\begin{tabular}{cccc|ccc}
\toprule  
Semantics 
 &\multicolumn{3}{c}{AWA1} &\multicolumn{3}{c}{AWA2}\\
  &u & s &H  & u & s &H \\

\midrule  
Attribute & 64.7 &81.1& 72.0& 76.8 & 82.7 & 79.6\\
Word2vec & 61.6 & 80.0 &69.5 & 71.4 &81.9 & 76.3\\
BERT & \textbf{77.9}& \textbf{81.8}& \textbf{79.8}& \textbf{79.0}& \textbf{86.7} &\textbf{82.7} \\

\bottomrule
\end{tabular}}

\end{center}
\end{table}

\subsubsection{\textbf{Robustness of Student Network}}
We elucidate the robustness inherent to the student network in this section. Given that the teacher network remains undisclosed by the Data Owner within the black-box protocol, it becomes imperative to showcase the results across diverse student models in this black-box scenario. As illustrated in Table \ref{BB-Snet}, the performances across various student models are closely aligned, denoting the stability and consistency afforded by our method.

\begin{table}[!tb]
\centering
\caption{Results with different student models in black-box protocol with omniscient teacher in GZSL task.}
\label{BB-Snet}
\renewcommand\arraystretch{0.7}
\small
\resizebox{0.45\textwidth}{!}
{
\begin{tabular}{cccc|ccc}
\toprule  
\textbf{Student} 
 &\multicolumn{3}{c}{\textbf{AWA1}} &\multicolumn{3}{c}{\textbf{aPY}}\\
 \textbf{Model} &u & s &H  & u & s &H \\

\midrule  
1 hidden layer & 31.9 & 25.9 & 28.6 & \textbf{30.4} & 34.4& 32.3\\
3 hidden layer & 27.9 & 26.3 & 27.1 & 28.5 & 36.4 & 32.0\\
Ours & \textbf{33.5} &\textbf{28.6} &\textbf{30.9} & 30.2& \textbf{42.2}& \textbf{35.2}\\

\bottomrule 
\end{tabular}
}
\end{table}
\subsection{\textbf{Potential Applications}}
As for potential applications, our SG-ZSL paradigm could carry profound implications for industries where data privacy is paramount. In healthcare, SG-ZSL can facilitate the sharing of medical insights without exposing patient data, thus advancing research while complying with stringent confidentiality regulations. Similarly, in finance, SG-ZSL enables the collaborative development of predictive models without risking sensitive financial information. Consequently, SG-ZSL fosters a collaborative environment where both data owners and AI service providers can thrive, leveraging the strengths of each party without compromising on security or copyright.

\subsection{\textbf{Limitations}} Although our research raises awareness of data and model privacy in the ZSL field, balancing privacy with performance remains challenging. The white-box protocol offers high performance through the guidance of teacher model weights and outputs but demands a careful balance between privacy and performance using differential privacy techniques. Meanwhile, the inherently secure black-box protocol may lag in optimization and performance due to its exclusive reliance on output-based supervision. Future efforts aim to bridge these gaps by enhancing the generator’s capabilities, notably by incorporating common-sense knowledge from large-scale models to establish a more robust knowledge space, thus improving knowledge transfer from seen to unseen classes.


\section{Conclusion}
\noindent In this work, we introduced an \ac{SG-ZSL} paradigm facilitating through data-free knowledge transfer. A pre-trained teacher model was instantiated at the data owner's end, acting as a data sentinel to render guidance for model training. A thorough evaluation was conducted for both `black-box' and `white-box' protocols, elucidating the trade-off between model performance and data privacy. Based on the proposed paradigm, the real data does not participate in the training at the AI service provider end, our model exhibits comparable performance against CZSL and GZSL while the data privacy is also secured. 
Future advancements in SG-ZSL can explore advanced optimization strategies based on more representative common knowledge (\ie from Large Language Models), and investigate more robust privacy protections, ensuring data owner interests are preserved without compromising model performance.


\section*{Acknowledgments}
This work is supported by the UK Medical Research Council (MRC) Innovation Fellowship under Grant MR/S003916/2, International Exchanges 2022 IEC$\backslash$NSFC$\backslash$223523.


\bibliographystyle{ieeetr}
\bibliography{sg-zsl}

\end{document}